\documentclass[10pt,twocolumn,letterpaper]{article}

\usepackage{cvpr}              

\usepackage{multirow}
\usepackage[most]{tcolorbox}
\usepackage{xcolor}

\definecolor{prompttitle}{HTML}{F6B26B} 
\definecolor{promptframe}{HTML}{F6B26B} 

\definecolor{frame}{HTML}{87CEFA}
\definecolor{title}{HTML}{87CEFA}

\definecolor{cvprblue}{rgb}{0.21,0.49,0.74}
\usepackage[pagebackref,breaklinks,colorlinks,allcolors=cvprblue]{hyperref}
\usepackage[table]{xcolor}
\def\paperID{41477} 
\def\confName{CVPR}
\def\confYear{2026}

\title{EvoGraph-R1: Self-Evolving Multimodal Knowledge Hypergraphs for Agentic Retrieval}

\author{
    Jiashi Lin\textsuperscript{1,2}\thanks{Equal contribution} \quad
    Changhong Jiang\textsuperscript{1}\thanks{Corresponding author} \quad
    Xiangru Lin\textsuperscript{3}\footnotemark[1] \quad
    Ruifei Zhang\textsuperscript{5} \quad
    Xinyi Zhu\textsuperscript{1} \\
    Jiyao Liu\textsuperscript{2} \quad
    Cheng Tang\textsuperscript{2} \quad
    Ye Du\textsuperscript{2} \quad
    Shujian Gao\textsuperscript{2} \quad
    Junzhi Ning\textsuperscript{2} \quad
    Lihao Liu\textsuperscript{2} \\
    Ziyan Huang\textsuperscript{2} \quad
    Tianbin Li\textsuperscript{2} \quad
    Jin Ye\textsuperscript{2,4} \quad
    Junjun He\textsuperscript{2} \\[3mm]
    \textsuperscript{1}Northwestern Polytechnical University,
    \textsuperscript{2}Shanghai Artificial Intelligence Laboratory \\
    \textsuperscript{3}The University of Hong Kong,
    \textsuperscript{4}Monash University,
    \textsuperscript{5}The Chinese University of Hong Kong (Shenzhen)
    \\[1mm]                                      {                                       \href{https://evograph-r1.github.io/}          {\texttt{https://evograph-r1.github.io/}}      }               
  }

\begin{document}
\maketitle
\begin{abstract}
Retrieval-augmented generation (RAG) has emerged as a critical paradigm for grounding Multimodal Large Language Models (MLLMs) in external knowledge. Recent GraphRAG methods introduce structured entity-relation graphs to improve retrieval and reasoning. However, they remain limited by treating knowledge graphs as static data structures built offline and queried in a single pass. This static paradigm misaligns with the interactive, iterative nature of knowledge-intensive reasoning, creating three bottlenecks: (i) text-centric fragmentation that impedes cross-modal reasoning, (ii) frozen structures unable to incorporate new evidence or correct errors, and (iii) rigid single-pass retrieval without adaptive refinement. To overcome these limitations, we introduce \textbf{EvoGraph-R1}, a self-evolving GraphRAG framework that reconceptualizes knowledge graphs as \textbf{dynamic environments} shaped through agent interactions.
We formulate retrieval as a Markov Decision Process (MDP) where the agent observes the graph state and executes actions to query (\textsc{GraphRetrieve}), expand (\textsc{WebSearch}), refine (\textsc{GraphEdit}), or terminate (\textsc{Answer}) the reasoning. These actions reshape the hypergraph structure and generate feedback signals that guide subsequent evolution.
Through this closed loop, the hypergraph evolves by integrating new evidence, correcting errors, and refining structure to support multi-hop reasoning. 
Experiments on multimodal VQA and text QA benchmarks demonstrate substantial improvements over existing RAG baselines in accuracy, coverage, and traceability, establishing self-evolving knowledge graphs as a fundamental paradigm across modalities.
\end{abstract}    
\section{Introduction}


\begin{figure}[htbp]
    \centering
    \includegraphics[width=0.95\linewidth]{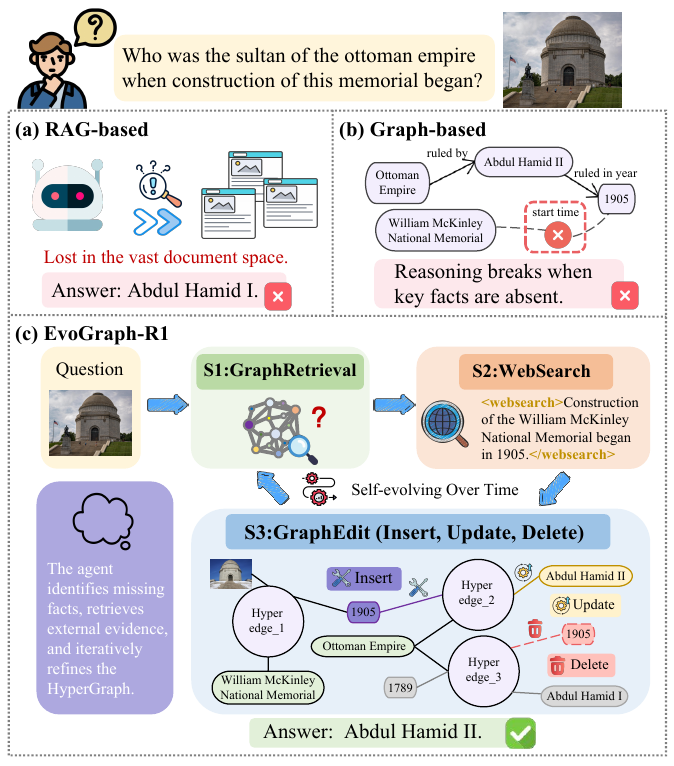}
    \caption{\textbf{Comparison of Retrieval Paradigms.} (a) Vanilla RAG loses information in document spaces. (b) Static graph-based RAG breaks when key facts are missing. (c) EvoGraph-R1 self-evolves through agent-driven \textsc{GraphRetrieve}, \textsc{WebSearch}, and \textsc{GraphEdit} to complete reasoning chains.}
    \label{fig:intro}
\end{figure}

Multimodal large language models (MLLMs) excel at visual reasoning tasks \cite{achiam2023gpt,liu2024llava}, yet their parametric memory remains limited \cite{mallen2023not}. No model can internalize the vast, ever-evolving knowledge of the open world. Retrieval-augmented generation (RAG) addresses this by enabling dynamic access to external knowledge at inference time, improving factual grounding in knowledge-intensive visual question answering (VQA) \cite{hu2023reveal}. However, existing multimodal RAG frameworks struggle with complex compositional reasoning \cite{ke2025explain}, as retrieved evidence is often reduced to disjoint text snippets or coarse page-level screenshots \cite{tanaka2025vdocrag,faysse2024colpali,mace2025vidore}, hindering fine-grained structural modeling and cross-modal alignment.
 
Recent GraphRAG methods~\citep{edge2024local,guo2024lightrag,guo2025rag,luo2025hypergraphrag} mitigate these limitations by leveraging knowledge graphs for retrieval and reasoning, thereby facilitating structured reasoning. The typical pipeline comprises three stages: (i) constructing graphs via LLM-based entity-relation extraction \cite{ye2022generative}; (ii) retrieving relevant subgraphs; and (iii) generating answers conditioned on graph knowledge. However, we identify a key limitation: \textbf{existing systems treat the knowledge graph as a static structure}, built offline in a single pass. This one-shot process makes graph quality difficult to guarantee, especially on long or multimodal inputs. As illustrated in Figure~\ref{fig:intro}, such static pipelines often yield incomplete or noisy graphs that propagate errors throughout retrieval and reasoning. Consequently, these limitations manifest as three bottlenecks: \textbf{(1) Text-centric fragmentation.} LLM-based extraction collapses rich multimodal evidence into isolated textual tuples, leading to missing relations, fragmented structures, and loss of long-range or cross-modal dependencies \cite{wan2025mmgraphrag}. \textbf{(2) Static structure.} Graphs are constructed offline and remain fixed during inference, preventing the incorporation of new evidence, error correction, or coverage of unseen topics \cite{pan2023large}. \textbf{(3) Rigid retrieval.} Retrieval is executed as a single-pass operation without iterative refinement. When initial evidence is insufficient, the system cannot revise its strategy, explore alternatives, or invoke external search \cite{trivedi2023interleaving}.

These limitations stem from treating the knowledge graph as a \textit{passive data structure} rather than an \textit{active reasoning substrate}. Human reasoning over knowledge is inherently interactive and iterative \cite{asaiself}: we collect initial evidence, identify gaps, seek additional information, reconcile contradictions, and progressively refine our understanding. This mirrors the agent-environment interaction paradigm in reinforcement learning \cite{luo2025graph}, where agents perceive states, take actions that modify those states, and receive feedback to guide future decisions.
Inspired by this paradigm, we reconceptualize multimodal GraphRAG as a \textbf{Markov Decision Process (MDP)} \cite{leng2025decex}, where the knowledge graph becomes a dynamic environment that co-evolves with the reasoning process:
\begin{itemize}[leftmargin=*,nosep]
\item \textbf{Environment State}: the multimodal knowledge hypergraph, encoding retrieved evidence (entities, relations, visual objects) and their structural organization (connectivity, alignment, relevance scores).
\item \textbf{Actions}: operations that \textit{query} the graph (\textsc{GraphRetrieve}), \textit{expand} it (\textsc{WebSearch}) for external knowledge), \textit{refine} it (\textsc{GraphEdit} to insert/update/delete elements), or \textit{terminate} (\textsc{Answer}).
\item \textbf{Rewards}: trajectory-level signals that assess reasoning quality (structural correctness and answer accuracy) and efficiency (computational cost).
\item \textbf{Policy}: the agent's action selection strategy, learned through reinforcement learning to maximize answer accuracy while maintaining efficiency.
\end{itemize}
This formulation enables a \textbf{self-evolving paradigm} \cite{cao2025knowledge}: the agent does not merely retrieve from a fixed graph but actively shapes it by adding verified evidence, correcting inconsistencies, pruning noise, and reorganizing structures to support multi-hop inference.

To instantiate this paradigm, we present \textbf{EvoGraph-R1}, an agentic multimodal GraphRAG framework that unifies retrieval and knowledge evolution into a closed-loop interaction process. Initialized from a minimal hypergraph, EvoGraph-R1 progressively refines its structure through agent–environment interactions. At each step, the agent observes the current graph state to identify knowledge gaps, then selects an action
that reshapes the hypergraph and generates feedback signals to guide subsequent decisions. Through this iterative cycle, EvoGraph-R1 transforms an initially sparse, noisy hypergraph into a coherent, query-specific knowledge state. 
Our self-evolving MDP paradigm generalizes beyond multimodal settings. By removing visual components while retaining the agent-driven evolution mechanism, the framework naturally adapts to text-only scenarios. We demonstrate state-of-the-art performance on both multimodal \cite{schwenk2022okvqa,huang2025frames} and text QA benchmarks \cite{yang2018hotpotqa}, validating that treating knowledge graphs as interactive environments provides value independent of modality.
Our main contributions are as follows:
\begin{itemize}
    \item \textbf{Self-evolving retrieval paradigm.} We introduce the first framework that models multimodal knowledge graphs as MDP environments, unifying retrieval, reasoning, and knowledge evolution through agent-environment interaction.

    \item \textbf{EvoGraph-R1 framework.} We design an autonomous agent operating over multimodal knowledge hypergraphs via four action types (\textsc{GraphRetrieve}, \textsc{WebSearch}, \textsc{GraphEdit}, \textsc{Answer}), implementing closed-loop co-evolution of graph structure and reasoning.

    \item \textbf{State-of-the-art performance.} We conduct multiple experiments on knowledge-intensive multimodal VQA and text QA benchmarks. Experimental results show our framework substantially outperforms existing RAG, GraphRAG, and search-augmented baselines in accuracy, efficiency, and evidential traceability.

\end{itemize}

\section{Related Work}

\textbf{RAG and GraphRAG.} 
Retrieval-augmented generation enhances LLM factuality by retrieving external evidence during inference~\cite{lewis2020retrieval}. While standard RAG retrieves flat text chunks without structured reasoning~\cite{peng2024graph}, GraphRAG~\cite{edge2024local} represents knowledge as entity-relation graphs, retrieving subgraphs and reasoning paths~\cite{jiang2024ragraph,edge2024local}. Graph-R1~\cite{luo2025graph} frames retrieval as interactive agent-environment interaction over hypergraphs, optimizing multi-turn exploration with reinforcement learning~\cite{luo2025graph,gettler2016becoming}.
Despite these advances, most GraphRAG systems assume offline, fixed graphs with one-shot retrieval. Graphs cannot be updated with new evidence or repaired at inference time, limiting robustness in open-world settings. EvoGraph-R1 departs from this static paradigm by enabling online graph expansion, updating, and pruning through agent-driven evolution.

\textbf{Multimodal RAG.}
Multimodal RAG extends text-only retrieval across modalities through modality-specific pipelines: document-oriented systems preserve layouts as images~\cite{xu2020layoutlm,wang2024mineru,faysse2024colpali} but lack relational grounding; scene-graph approaches link visual regions to text but flatten structured elements~\cite{wan2025mmgraphrag}. VLM-based methods~\cite{mace2025vidore,tanaka2025vdocrag,sun2025visrag} retrieve document images to avoid OCR errors but introduce redundancy. RAG-Anything~\cite{guo2025rag} constructs graphs encoding cross-modal relationships, yet still assumes static, offline structures.
EvoGraph-R1 incrementally builds, updates, and prunes multimodal hypergraphs during inference, shifting retrieval from a static one-shot answering paradigm to maintaining a persistent, evolving knowledge state.

\textbf{Web-search Equipped MLLMs.} 
Recent work equips MLLMs with active web access for autonomous search and result integration. Search-R1~\cite{jin2025search} uses RL to interleave reasoning with web search; MMSearch-R1~\cite{wu2025mmsearch} extends this to multimodal settings with end-to-end RL. While demonstrating the value of interactive retrieval, these systems treat retrieved content as transient prompt context. EvoGraph-R1 instead regards each retrieval step, including web evidence, as a state update to a persistent hypergraph that is continuously expanded, revised, and filtered, thereby enabling systematic multi-hop reasoning with explicit traceability.
\section{Method}

\begin{figure}[htbp]
    \centering
    \includegraphics[width=1.0\linewidth]{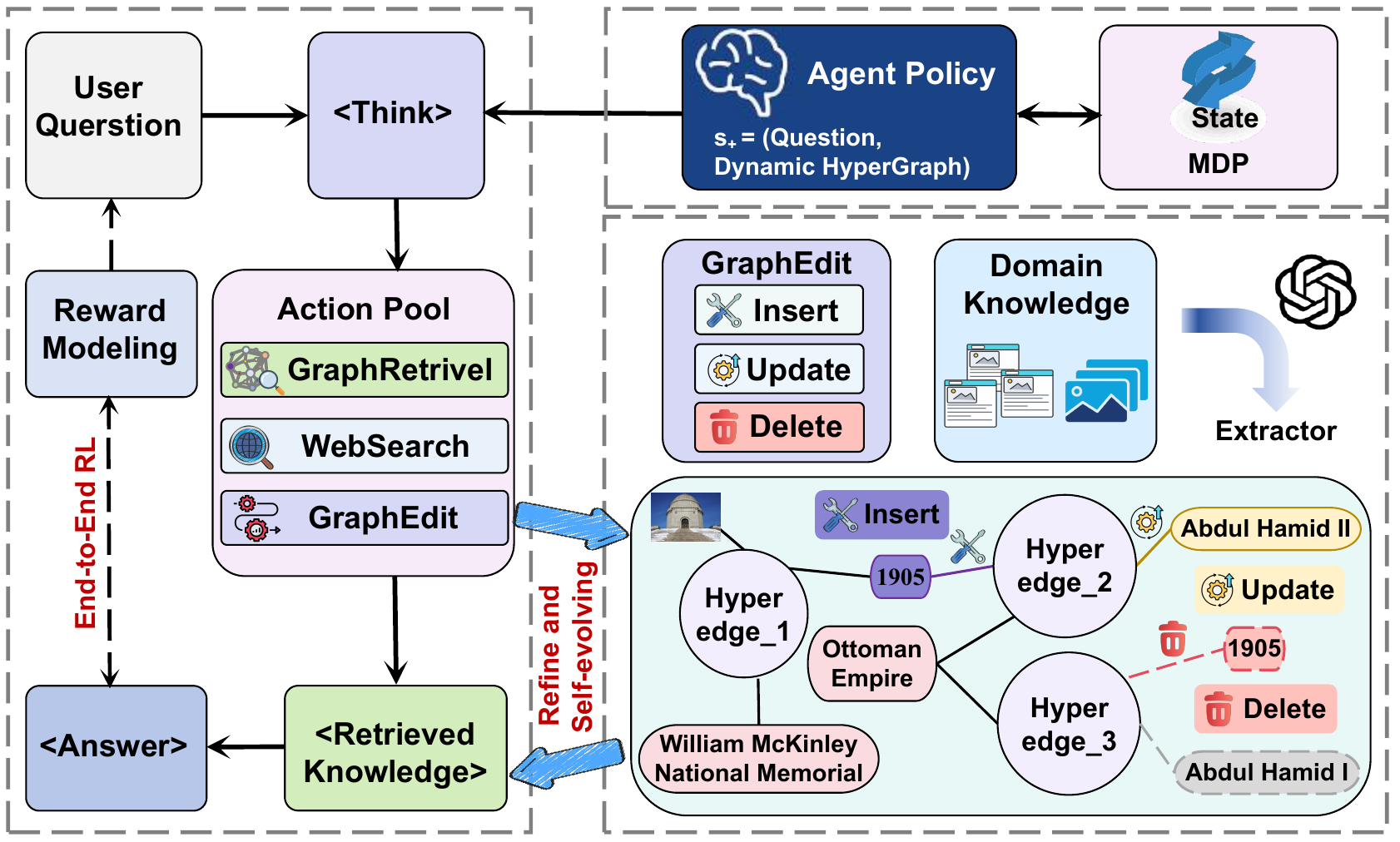}
   \caption{\textbf{EvoGraph-R1 framework overview.} The agent operates over a dynamic hypergraph formulated as an MDP. At each step, it selects actions \textsc{GraphRetrieve}, \textsc{WebSearch}, \textsc{GraphEdit}, or \textsc{Answer}. \textsc{GraphEdit} includes \textsc{Insert}, \textsc{Update}, and \textsc{Delete}, enabling continuous refinement. The agent policy is optimized via end-to-end RL with trajectory-level rewards, forming a closed-loop system in which the hypergraph co-evolves with the reasoning process.}

    \label{fig:method}
\end{figure}

\subsection{Overview}
EvoGraph-R1 comprises three components: (1) \textbf{Multimodal Hypergraph Construction} creates a unified cross-modal knowledge representation; (2) \textbf{Agent-Based Graph Evolution} implements iterative graph refinement through four actions; (3) \textbf{Reinforcement Learning} optimizes the agent policy using trajectory-level rewards. An overview is presented in Figure~\ref{fig:method}. 

\subsection{Problem Setup}
\label{sec:problem}
In knowledge-driven VQA, the model receives an image-question pair $(I_q, q)$ and generates an answer $a$ by retrieving relevant knowledge from $\mathcal{B} = \{(T_i, I_i)\}_{i=1}^{N}$, where each document contains text $T_i$ and image $I_i$. This requires visual and textual reasoning, demanding that the model align and synthesize cross-modal content effectively. Unlike traditional RAG systems that retrieve static documents, our framework maintains a persistent knowledge hypergraph that is incrementally expanded, updated, and pruned in response to queries, enabling systematic reasoning over interconnected multimodal evidence.

\subsection{Multimodal Hypergraph Construction}
\label{sec:hypergraph}
We construct a unified cross-modal knowledge hypergraph $\mathcal{G}_H = (\mathcal{V}, \mathcal{E})$ from a multimodal corpus through three stages: textual extraction, visual construction, and cross-modal fusion. This hypergraph serves as the initial environment state for agent-based retrieval and reasoning.

\textbf{Textual Subgraph Extraction}. Given the input text $d$, we segment it into knowledge fragments represented as hyperedges $\mathcal{E}_H^{(d)} = \{e_1, \ldots, e_k\}$, where each $e_i = (e_i^{\text{text}}, \mathcal{V}_{e_i}, r_i, \sigma_i)$ contains a natural-language description, entity set, relation type, and confidence score $\sigma_i \in (0, 10]$. Following prior work on hypergraph-structured knowledge~\cite{guo2024lightrag,luo2025hypergraphrag}, we employ an $n$-ary relation extraction prompt with an MLLM-based extractor $\pi_{\text{ext}}$:
\begin{equation}
\mathcal{F}_d^{(n)} = \{f_1, \ldots, f_k\} \sim \pi_{\text{ext}}(d),
\end{equation}
where each extracted $f_\ell = (e_\ell, \mathcal{V}_{e_\ell})$ associates a hyperedge $e_\ell$ with its participating entity set $\mathcal{V}_{e_\ell}$. This formulation allows each text segment to contribute multiple, potentially overlapping $n$-ary relational facts to the unified textual hypergraph $\mathcal{G}_H^{\text{text}} = (\mathcal{V}^{\text{text}}, \mathcal{E}^{\text{text}})$, facilitating high-order relational reasoning beyond binary relations.

\textbf{Visual Subgraph Construction}. While text naturally lends itself to entity-relation extraction, visual content requires special handling to preserve scene-level semantics. For each image $x \in I_i$, we employ extractor $\pi_{\text{ext}}$ to generate a detailed natural language scene description together with the primary object names detected within the image. This scene description is then encoded as a high-order hyperedge, serving as an ``anchor'' for all entities and relations grounded in the image:
\begin{equation}
u_x = (\text{id} = x, \text{type} = \text{image}, \text{mod} = \text{vis}),
\end{equation}
where $u_x$ is the anchor node representing the image. Entities and finer-grained visual relational facts are subsequently extracted from the scene description and associated with $u_x$, yielding
\begin{equation}
(\mathcal{V}_x^{\text{vis}}, \mathcal{E}_x^{\text{vis}}) = \pi_{\text{ext}}(x).
\end{equation}
Each visual hyperedge $e = (\mathcal{V}_e, r_e, \sigma_e, \text{mod} = \text{vis})$ includes the anchor node ($u_x \in \mathcal{V}_e$) ensuring all extracted information remains grounded in its image context, preventing visual facts from floating without source attribution. All image-specific subgraphs $\mathcal{G}_x^{\text{vis}} = (\{u_x\} \cup \mathcal{V}_x^{\text{vis}}, \mathcal{E}_x^{\text{vis}})$ are integrated across the corpus, forming the complete visual subgraph $\mathcal{G}_H^{\text{vis}} = (\mathcal{V}^{\text{vis}}, \mathcal{E}^{\text{vis}})$.

\textbf{Cross-Modal Graph Fusion}. To unify textual and visual subgraphs, we normalize entity labels and apply the entity-resolution function $\phi$, matching visual and textual mentions using string similarity and embedding proximity. Matched entities merge into canonical nodes, inheriting attributes and contextual edges from both modalities. All hyperedges from original subgraphs reconnect to merged nodes, yielding the fused graph $\mathcal{G}_H = (\mathcal{V}, \mathcal{E})$:
\begin{equation}
\mathcal{V} = \text{Resolve}(\mathcal{V}^{\text{text}} \cup \mathcal{V}^{\text{vis}}, \phi), \quad \mathcal{E} = \mathcal{E}^{\text{text}} \cup \mathcal{E}^{\text{vis}}.
\end{equation}

\textbf{Offline Graph Indexing}. For efficient retrieval during agent interaction, we embed all graph elements (textual entities, visual entities, hyperedges) using multimodal encoder $f(\cdot)$ (e.g., GME~\cite{zhang2024gme}) into a shared semantic space:
\begin{equation}
\phi_{\text{entity}} = f(v^{\text{text}} \cup v^{\text{vis}}), \quad \phi_{\text{edge}} = f(e), \; \forall e \in \mathcal{E}.
\end{equation}
This unified hypergraph $\mathcal{G}_H = (\mathcal{V}, \mathcal{E})$ retains fine-grained cross-modal semantics and forms the initial environment state for agent-driven evolution.

\subsection{Agent-Based Graph Evolution}
\label{sec:agent}
While the initial hypergraph $\mathcal{G}_H$ provides structured knowledge, automatically constructed graphs are often incomplete and noisy. Moreover, static graphs fundamentally cannot adapt to query-specific information needs or incorporate newly discovered evidence. To address these limitations, we formulate retrieval as an agent-environment interaction process where the knowledge graph becomes a dynamic environment that co-evolves with reasoning. 

\subsubsection{MDP Formulation}
We model the agent-graph interaction as a discrete-time Markov decision process over steps $t = 0, 1, \ldots, T$. At initialization, the agent receives the multimodal hypergraph $\mathcal{G}_0$ and query $q$. The process terminates when the agent selects \textsc{Answer} action or reaches step limit $T$, then it generates a final answer conditioned on the evolved hypergraph $\mathcal{G}_T$. We formalize the key components of this MDP as:

\textbf{Environment State.} At step $t$, the environment state captures the evolving knowledge context:
\begin{equation}
s_t = (\mathcal{G}_t, \mathcal{H}_t, q),
\end{equation}
where $\mathcal{G}_t = (\mathcal{V}_t, \mathcal{E}_t)$ is the current hypergraph encoding retrieved evidence (entities, relations, visual regions) and their structural organization (connectivity, cross-modal alignments, relevance scores), $\mathcal{H}_t$ is the action history (a sequence of tuples $(a_k, r_k, \mathcal{G}_{k+1})$ for $k < t$), and $q$ is the input query. This representation ensures the agent conditions its behavior on both current graph structure and reasoning trajectory, enabling coherent multi-turn reasoning and credit assignment.

\textbf{Actions.} The agent operates over a discrete action space $\mathcal{A}$ with four action types:

\textit{(1) \textsc{GraphRetrieve}} queries the current environment via elements in hypergraph $\mathcal{G}_t$. It supports cross-modal entity-level lookups over surrounding subgraphs and hyperedge-level vector retrieval, enabling the agent to exploit structured, semantically linked higher-order facts.

\textit{(2) \textsc{WebSearch}} is invoked when in-graph evidence proves insufficient. The agent formulates a query $q_{\text{web}}$ and retrieves external information from web search engines. The returned content is aligned with the current state and appended to the textual context for subsequent generation, allowing the agent to incorporate external evidence beyond the static knowledge base.

\textit{(3) \textsc{GraphEdit}} refines the environment through direct graph-structure modification via three complementary sub-operations: \textsc{Insert}, which introduces new entities or hyperedges derived from retrieved content and expands the graph with verified evidence; \textsc{Update}, which revises existing hyperedges to correct errors, resolve conflicts, or strengthen factual grounding based on newly discovered information; and \textsc{Delete}, which performs soft removal by lowering the confidence scores of low-quality or contradicted elements, thereby pruning noise.
Together, these operations enable self-correction and adaptive refinement of the knowledge graph as reasoning progresses, effectively addressing the inherent incompleteness and noise of automatically constructed graphs.

\textit{(4) \textsc{Answer}} terminates the reasoning process. Once the agent determines that the environment has evolved sufficiently and accumulated adequate evidence, it invokes this action to produce the final response based on the current graph state $\mathcal{G}_t$.

\textbf{Rewards.} After each action, the agent receives reward signals based on three factors: (1) whether reasoning steps follow the intended interaction protocol (structural correctness), (2) semantic alignment between predicted and ground-truth answers (answer quality), and (3) the computational cost of executed actions (efficiency). These rewards guide policy learning to maximize answer accuracy while maintaining efficient graph evolution (Section~\ref{sec:rl}).

\textbf{Policy.} The agent's policy $\pi_\theta(a_t \mid s_t)$ determines action selection based on current state, learned through reinforcement learning (Section~\ref{sec:rl}) to maximize answer accuracy while maintaining efficiency.

\subsubsection{MDP State Transition Dynamics}
The hypergraph is dynamic and responsive to agent actions, enabling self-evolution. At each step $t$, action $a_t$ induces a transition via evolution policy $\pi_{\text{evolve}}$:
\begin{equation}
\mathcal{G}_{t+1} = \pi_{\text{evolve}}(a_t, \mathcal{G}_t),
\end{equation}
where different action types induce different state changes:
\begin{itemize}[leftmargin=*,nosep]
\item \textsc{GraphRetrieve} marks which entities and relations have been accessed without altering structure.
\item \textsc{WebSearch} acts as an external interface, retrieving evidence to validate and augment the graph with new facts and cross-modal alignments.
\item \textsc{GraphEdit} operations directly refine graph quality by consolidating fragmented knowledge, correcting inconsistencies, and removing noise.
\end{itemize}

\noindent Each transition is appended to the action history:
\begin{equation}
\mathcal{H}_{t+1} = \mathcal{H}_t \cup (a_t, r_t, \mathcal{G}_{t+1}),
\end{equation}
forming a cumulative record that guides online decision-making and provides supervision signals for reinforcement learning. The final output is computed as:
\begin{equation}
a^* \leftarrow \pi_{\text{resp}}^{\text{final}}(q, \mathcal{H}_T).
\end{equation}

Through this recurrent update mechanism, the agent gradually transforms an initially sparse, noisy hypergraph $\mathcal{G}_0$ into a coherent, query-specific knowledge state $\mathcal{G}_T$ optimized for reasoning. This evolutionary process yields three benefits that address the limitations of static GraphRAG: (1) fine-grained multimodal entities and cross-modal alignments are dynamically constructed rather than pre-defined, overcoming text-centric fragmentation; (2) the graph continuously evolves to integrate new evidence and correct errors rather than remaining frozen, eliminating static structure constraints; (3) retrieval adapts through multi-turn exploration guided by intermediate reasoning rather than single-pass lookup, removing rigid pipeline limitations.

\subsection{Reinforcement Learning Optimization}
\label{sec:rl}

We train the agent's policy to efficiently evolve the knowledge hypergraph using Group Relative Policy Optimization (GRPO) ~\cite{shao2024deepseekmath} with a trajectory-level reward function that balances answer correctness, reasoning coherence, and retrieval efficiency.

\subsubsection{Training Setup}

Given a dataset of questions $\mathcal{D}_Q$, we train the agent through experience collected from its interactions with the hypergraph environment. For each question $q \in \mathcal{D}_Q$, an initial state $s_1 \sim P(q)$ is sampled (consisting of the query $q$, a relevant initial subgraph, and an empty history). At step $t$, the agent observes $(s_{t-1}, \mathcal{G}_H)$ and selects an action $a_t \sim \pi_\theta(\cdot \mid s_{t-1}; \mathcal{G}_H)$ according to the current policy parameterized by $\theta$. If $a_t$ is an \textsc{GraphEdit} action, the agent also proposes specific modifications $m'_t$ to commit to $\mathcal{G}_H$. A complete trajectory is defined as:
\begin{equation}
\tau = (s_1, a_1, s_2, a_2, \ldots, s_{|\tau|}) \in \mathcal{T}_q,
\end{equation}
where only \textsc{GraphEdit} actions modify the graph state $\mathcal{G}_H$, while other actions (e.g., \textsc{GraphRetrieve}, \textsc{WebSearch}) gather information without altering structure.

\subsubsection{Reward Design and Policy Optimization}

To optimize both reasoning quality and graph evolution efficiency, we define a trajectory-level reward function $R(\tau)$ composed of three complementary components:

\textbf{(i) Structural Reward} evaluates whether each reasoning step follows the intended interaction protocol. At each step $(s_t, a_t)$, the agent produces an intermediate block $(a^{\text{think}}_t, \alpha_t, a^{\text{out}}_t)$. Well-formed steps are rewarded, up to a capped total:
\begin{equation}
R_{\text{struct}}(\tau) = \min\left(1.0, \eta \sum_{t=1}^{|\tau|} \mathbb{I}_{\text{valid}}(t)\right),
\end{equation}
where $\mathbb{I}[\cdot]$ denotes the indicator function, and $\eta$ is a step-wise scaling factor set to 0.5. This component encourages disciplined reasoning and syntactic consistency across multi-turn trajectories.

\textbf{(ii) Answer Reward} measures semantic fidelity using a token-level overlap between the predicted answer and the ground-truth label $y^*_q$:
\begin{equation}
R_{\text{ans}}(a^{\text{ans}}_T) = \frac{2 |\text{tok}(a^{\text{ans}}_T) \cap \text{tok}(y^*_q)|}{|\text{tok}(a^{\text{ans}}_T)| + |\text{tok}(y^*_q)|},
\end{equation}
where $\text{tok}(\cdot)$ applies normalization such as lowercasing and whitespace removal. This F1-style score rewards semantic alignment without requiring exact string matching.

\textbf{(iii) Overall Outcome Reward} integrates correctness, structure, and efficiency:

\begin{equation}
\begin{aligned}
R(\tau) = &-\lambda \sum_{t=1}^{|\tau|} c(a_t) + R_{\text{struct}}(\tau) \\
&+ \mathbb{I}[R_{\text{struct}}(\tau) = 1.0] \cdot R_{\text{ans}}(a^{\text{ans}}_T),
\end{aligned}
\end{equation}
where $c(a_t)$ denotes the normalized cost of executing action $a_t$ (e.g., retrieval or graph edit), and $\lambda > 0$ is the efficiency coefficient. This formulation ensures correctness is rewarded only when the reasoning process remains structurally coherent, thereby promoting high-quality, cost-efficient graph evolution.

\textbf{(iv) Policy Optimization}. We optimize the policy using GRPO, which computes group-level advantages by comparing each trajectory's reward to the group mean. This enables stable updates and efficient learning from multiple rollouts per question. Through this optimization, the agent learns to navigate the action space, selectively retrieving and editing graph elements to construct query-specific knowledge states that enable accurate question answering.
\section{Experiments}
We conduct a comprehensive evaluation of EvoGraph-R1 to address the following questions: \textbf{\emph{RQ1:}} Does EvoGraph-R1 outperform existing baselines in reasoning accuracy? \textbf{\emph{RQ2:}} To what extent do the individual components contribute to its effectiveness? \textbf{\emph{RQ3:}} What is the cost of constructing the unified heterogeneous hypergraph? \textbf{\emph{RQ4:}} How well does it generate accurate and coherent answers? Finally, \textbf{\emph{RQ5:}} how does it perform in low-resource settings?

\subsection{Experiment Setup}

\textbf{Datasets.} We evaluate on both \emph{multimodal} and \emph{text-only} settings to demonstrate the generalizability of our self-evolving paradigm. For multimodal tasks, we use E-VQA~\cite{huang2025frames}, InfoSeek~\cite{chen2023can}, and OK-VQA~\cite{marino2019ok}, with Wikipedia as the knowledge source filtered via EchoSight~\cite{hu2023reveal}. For text-only tasks, we use 2WikiMultiHopQA~\cite{ho2020constructing}, HotpotQA~\cite{yang2018hotpotqa}, and Natural Questions (NQ)~\cite{yang2018hotpotqa,lewis2020retrieval} with official Wikipedia dumps.

\textbf{Evaluation Metrics.} We use the LLM-as-Judge framework for multimodal accuracy, F1~\cite{jin2025flashrag} for text-only performance, and G-E~\cite{que2024hellobench} for answer quality.





\textbf{Baselines.} We compare against three categories: \emph{(i) Standard RAG methods}: NaiveRAG~\cite{lewis2020retrieval}, EchoSight~\cite{hu2023reveal}, ReflectiVA~\cite{sun2025visrag}, and MMKB-RAG; \emph{(ii) GraphRAG methods}: GraphRAG~\cite{edge2024local}, LightRAG~\cite{guo2024lightrag}, HippoRAG2, and HyperGraphRAG~\cite{luo2025hypergraphrag}; \emph{(iii) RL-augmented methods}: Search-R1~\cite{jin2025search}, MMSearch-R1~\cite{wu2025mmsearch} and Graph-R1~\cite{luo2025graph}.

\textbf{Implementation Details}. We employ GPT-4o-mini for knowledge construction in both the EvoGraph-R1 and GraphRAG baselines. For retrieval, we use GME~\cite{zhang2024gme}. EvoGraph-R1 is implemented with Qwen2.5-VL-7B as the base MLLM for multimodal tasks and Qwen2.5-7B-Instruct for text-only tasks. We conduct experiments using three random seeds and run all experiments on four 80GB NVIDIA A100 GPUs. Additional implementation details are provided in the \underline{supplementary document}.

\subsection{Comparative Performance Analysis - RQ1}
Tables~\ref{table:textonly} and~\ref{table:multimodal} present comprehensive comparisons across text-only and multimodal benchmarks. EvoGraph-R1 consistently outperforms all baselines, with improvements more pronounced on multimodal tasks.

\textbf{Text-Only Performance.} We outperform the strongest baseline (Graph-R1) by +3.5\% F1 on 2WikiMultiHopQA, +2.7\% on HotpotQA, and +6.9\% on Natural Questions, achieving 68.5\%, 65.4\%, and 56.8\% respectively. These improvements underscore the effectiveness of our dynamic graph evolution and retrieval framework. By progressively refining entity relations and evidence pathways, our method surpasses static text-based and graph-based approaches.

\textbf{Multimodal Performance.} On E-VQA, we achieve 43.6\% accuracy, outperforming MMSearch-R1 by +6.7\% and MMKB-RAG by +7.7\%. On OK-VQA, we reach 68.6\%, surpassing GPT-4o-mini by +2.7\%. These substantial improvements confirm that dynamically constructing and refining multimodal knowledge hypergraphs is more effective than retrieving static documents.


\textbf{Comparison with Web-search equipped Methods.} 
EvoGraph-R1 maintains a persistent, evolving knowledge state instead of discarding evidence after each step. This enables cumulative reasoning and structured integration of noisy web content. It outperforms MMSearch-R1 by +21.8\% and Search-R1 by +17.5\% on text-only benchmarks, while also surpassing MMSearch-R1 by +6.7\% on E-VQA and +8.7\% on OK-VQA, confirming the advantage of dynamic graph evolution over transient retrieval across both text-only and multimodal benchmarks. Importantly, EvoGraph-R1’s gains stem from the synergy between the persistent hypergraph environment and RL-driven policy evolution, which together provide both accuracy and efficiency across text-only and multimodal reasoning.

\begin{table}[htbp]
\centering
\footnotesize
\caption{\textbf{Performance comparison on text-only datasets.} \textbf{Bold} indicates best performance.}
\label{table:textonly}
\begin{tabular}{lcccc}
\toprule
\textbf{Method} & 2Wiki & HotpotQA & NQ & Avg. \\
\midrule
\multicolumn{5}{l}{\textit{Without Retrieval}} \\
Qwen2.5-7B-Instruct  & 12.2 & 16.6 & 13.2 & 14.00 \\
GPT-4o-mini & 17.0 & 31.8 & 21.6 & 23.47 \\
\midrule
\multicolumn{5}{l}{\textit{Retrieval-Augmented Models}} \\
NaiveRAG & 20.1 & 44.3 & 24.6 & 29.67\\
\midrule
\multicolumn{5}{l}{\textit{Graph Retrieval-Augmented Models}} \\
GraphRAG & 16.0 & 31.7 & 20.3 & 22.67 \\
LightRAG & 17.6 & 30.7 & 19.1 & 22.47 \\
HippoRAG2 & 23.5 & 44.8 & 24.1 & 30.80 \\
\midrule
\multicolumn{5}{l}{\textit{Retrieval-Augmented Models with Reinforcement Learning}} \\
Search-R1-7B & 41.3 & 50.9 & 45.9 & 46.03 \\
MMSearch-R1-7B & 38.5 & 45.3 & 41.6 & 41.80 \\
Graph-R1-7B & 65.0 & 62.7 & 49.9 & 59.20 \\
\midrule
\rowcolor{blue!10}
EvoGraph-R1-7B & \textbf{68.5} & \textbf{65.4} & \textbf{56.8} & \textbf{63.57} \\
\bottomrule
\end{tabular}
\vspace{-3mm}
\end{table}

\begin{table}[htbp]
\centering
\footnotesize
\caption{\textbf{Performance comparison on multimodal datasets.} \textbf{Bold} indicates best performance.}
\label{table:multimodal}
\begin{tabular}{lcccc}
\toprule
\textbf{Method} & E-VQA & InfoSeek & OK-VQA & Avg. \\
\midrule
\multicolumn{5}{l}{\textit{Without Retrieval}} \\
Qwen2.5-VL-7B & 18.8 & 26.4 & 35.0 & 26.73 \\
GPT-4o-mini & 27.2 & 35.9 & 65.9 & 43.00 \\
\midrule
\multicolumn{5}{l}{\textit{Retrieval-Augmented Models}} \\
NaiveRAG & 27.4 & 28.4 & 56.1 & 37.30 \\
Echosight & 21.7 & 30.4 & -- & 26.05 \\
ReflectiVA & 29.2 & 40.1 & -- & 34.65 \\
MMKB-RAG & 35.9 & 36.4 & 65.4 & 45.90 \\
\midrule
\multicolumn{5}{l}{\textit{Graph Retrieval-Augmented Models}} \\
GraphRAG & 19.8 & 24.3 & 51.8 & 31.97 \\
LightRAG & 17.7 & 20.3 & 46.3 & 28.10 \\
HippoRAG2 & 22.6 & 25.6 & 48.2 & 32.13 \\
\midrule
\multicolumn{5}{l}{\textit{Retrieval-Augmented Models with Reinforcement Learning}} \\
MMSearch-R1-7B & 36.9 & 41.3 & 59.9 & 46.03 \\
Graph-R1-7B & 28.5 & 29.8 & 53.6 & 37.30 \\
\midrule
\rowcolor{blue!10}
EvoGraph-R1-7B & \textbf{43.6} & \textbf{42.3} & \textbf{68.6} & \textbf{51.50} \\
\bottomrule
\end{tabular}
\vspace{-3mm}
\end{table}

\subsection{Ablation Study - RQ2}

Table~\ref{table:ablation} quantifies individual component contributions on 2WikiMultiHopQA and E-VQA.

\textbf{Multimodal Hypergraph Environment.} Removing the multimodal hypergraph causes a 5.4\% drop on 2Wiki F1 and 4.8\% drop on E‑VQA accuracy, with retrieval rounds increasing. This underscores its role in enabling structured reasoning and cross‑modal alignment through unified textual–visual representation.

\textbf{Graph Edit Operations.} \textsc{INSERT} is most critical: removing it causes an 8.4\% drop on 2Wiki and a 6.8\% drop on E-VQA, with 0.8 additional inefficient retrieval rounds. \textsc{UPDATE} provides substantial gains: removing it results in drops of 5.5\% and 3.9\% on 2Wiki and E-VQA, respectively, confirming the benefits of error correction. \textsc{DELETE} has a smaller but consistent impact, with 2.4\% and 1.5\% drops when removed, validating the benefits of noise filtering.

\textbf{Web Search.} Removing \textsc{WebSearch} leads to substantial performance drops of 9.6\% on 2Wiki and 11.2\% on E-VQA, revealing the limitations of static corpora in supporting open-domain and fine-grained multimodal reasoning, especially when handling long-tail knowledge. Moreover, one-shot graph construction exacerbates this limitation by introducing information loss and structural ambiguity, which disrupts the continuity of reasoning.



\begin{table}[t]
\centering
\footnotesize
\setlength{\tabcolsep}{5pt}
\caption{\textbf{Ablation study on 2WikiMulti-HopQA and E-VQA.} We report F1/Acc (\%) and average retrieval rounds as an efficiency indicator.}
\label{table:ablation}
\begin{tabular}{lcccc}
\toprule
\multirow{2}{*}{\textbf{Setting}} & \multicolumn{2}{c}{\textbf{2Wiki}} & \multicolumn{2}{c}{\textbf{E-VQA}} \\
\cmidrule(lr){2-3} \cmidrule(lr){4-5}
& F1.\,$\uparrow$ & Rounds\, & Acc.\,$\uparrow$ & Rounds\,\\
\midrule
\rowcolor{blue!10}
\textbf{Full (all components)}
    & \textbf{68.5} & \textbf{2.57}
    & \textbf{43.6} & \textbf{1.65} \\
\midrule
\textminus MM Hypergraph
    & 63.1 & 2.82
    & 38.8 & 2.29 \\
\textminus INSERT
    & 60.1 & 3.48
    & 36.8 & 2.45 \\
\textminus UPDATE
    & 63.0 & 2.95
    & 39.7 & 2.10 \\
\textminus DELETE
    & 66.1 & 2.70
    & 42.1 & 1.85 \\
\textminus WEBSEARCH
    & 58.9 & 3.17
    & 32.4 & 2.22 \\
\bottomrule
\end{tabular}
\vspace{-3mm}
\end{table}

\subsection{Retrieval Efficiency Analysis - RQ3}
We analyze the computational efficiency through retrieval rounds and response length. As shown in Table~\ref{table:ablation}, EvoGraph-R1 completes queries in 2.57 rounds on 2Wiki and 1.65 rounds on E-VQA, while ablated variants require significantly more: $-$\textsc{INSERT} needs 3.48 rounds, a +35.4\% increase, and $-$\textsc{WebSearch} requires 3.17 rounds, a +23.3\% increase, indicating inefficient searches for unavailable information.

\begin{figure}[h!]
	\centering
	\subfloat{\includegraphics[width=0.5\linewidth]{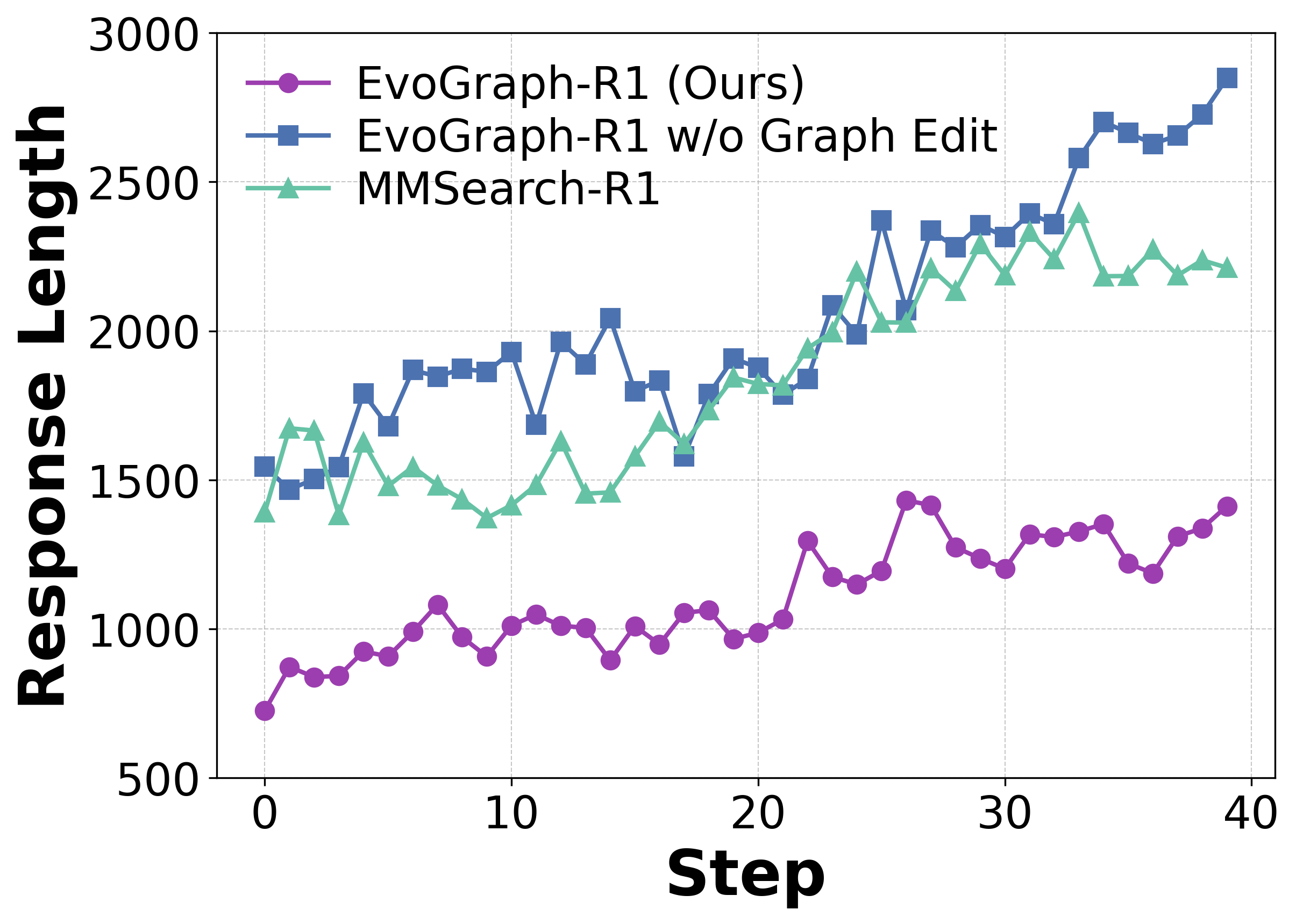}%
		\label{fig_sub1}}
	\subfloat{\includegraphics[width=0.5\linewidth]{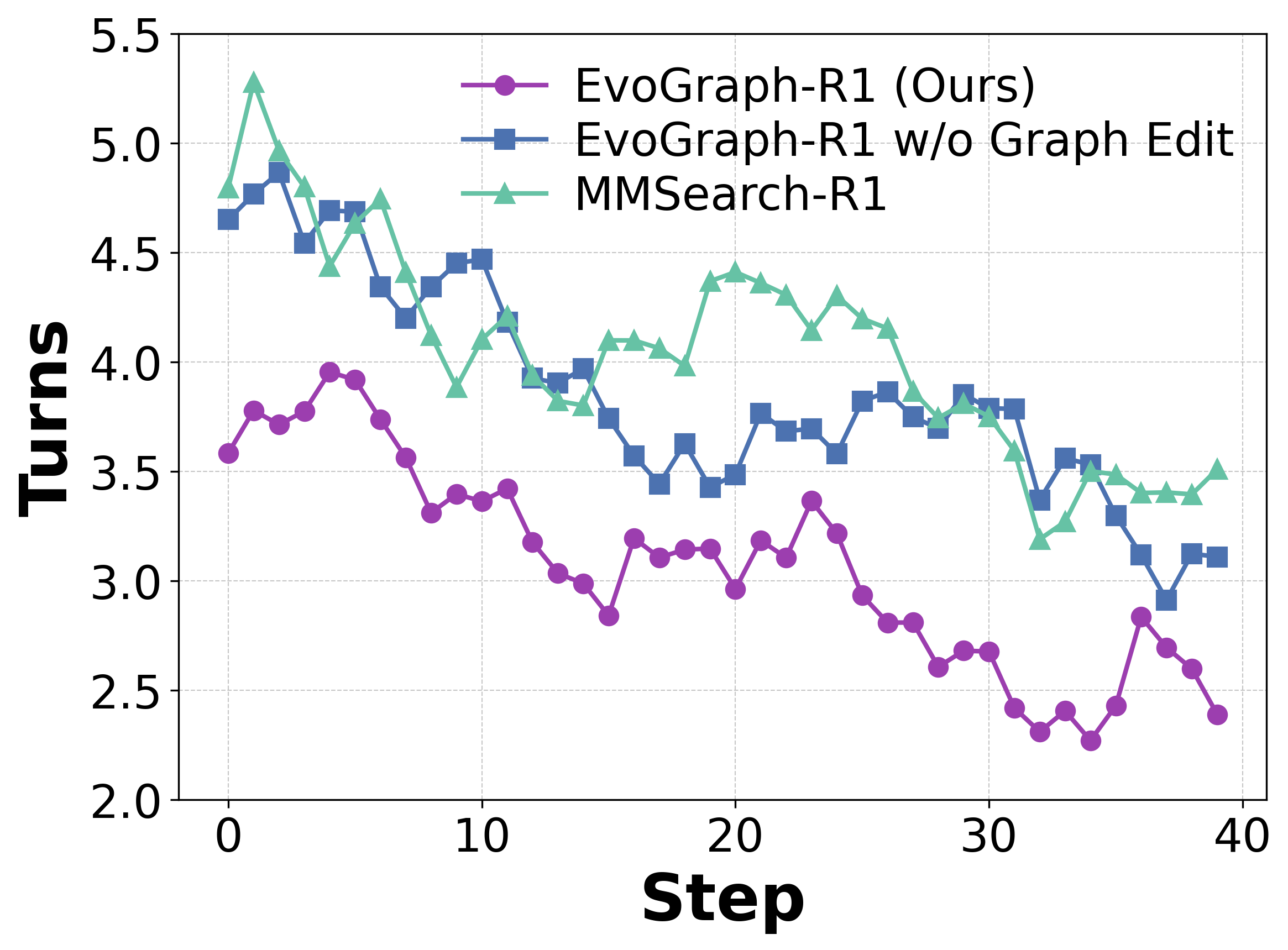}%
		\label{fig_sub2}}
	\caption{\textbf{Average Retrieval Turns and Response Length on 2Wiki.}}
	\label{fig:efficiency}
    \vspace{-3mm}
\end{figure}

Figure~\ref{fig:efficiency} further demonstrates this efficiency advantage. EvoGraph-R1, shown in purple, converges to fewer retrieval turns of approximately 2.4 and generates more concise responses of around 1,300 tokens. This compares favorably to variants without graph editing, shown in blue with approximately 3.1 turns and 2,850 tokens, and MMSearch-R1, shown in green with approximately 3.5 turns and 2,200 tokens. This dual efficiency gain validates that persistent hypergraph refinement enables targeted knowledge integration. The agent avoids redundant retrieval by maintaining a refined state and generates focused answers without compensatory verbosity.

\subsection{Generation Quality Evaluation - RQ4}

\begin{figure}[t]
    \centering
    \includegraphics[width=0.71\linewidth]{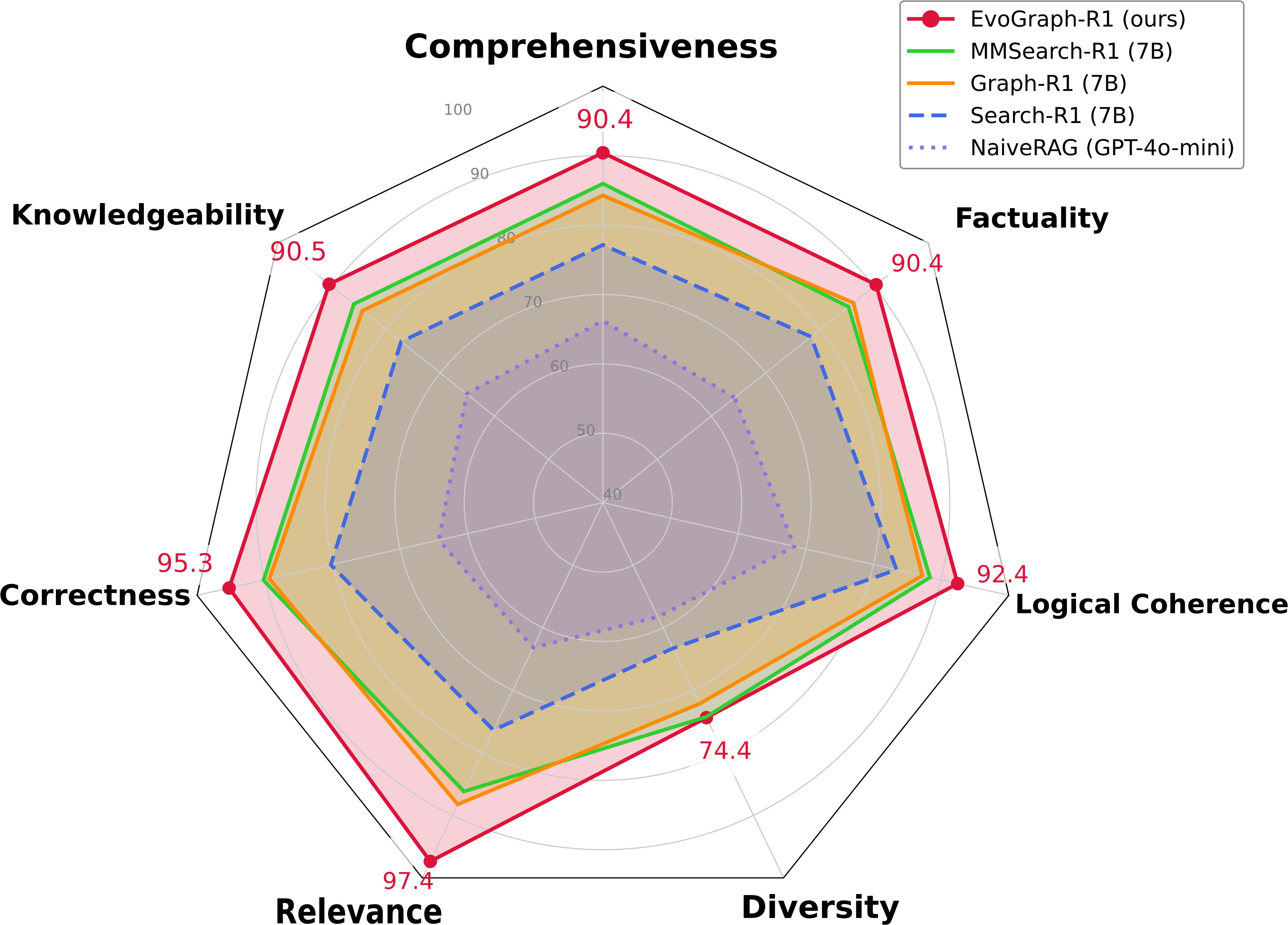}
    \caption{\textbf{Comprehensive quality comparison across seven evaluation dimensions.} EvoGraph-R1 (red) consistently outperforms all baselines, demonstrating superior generation quality through self-evolving graph refinement.}
    \label{fig:radar_comparison}
    \vspace{-1.8mm}
\end{figure}

Following HelloBench~\citep{que2024hellobench}, which employs GPT‑4o‑mini as LLM‑as‑Judge, we assess answer quality across seven dimensions. As shown in Figure~\ref{fig:radar_comparison}, EvoGraph‑R1 consistently surpasses all baselines, achieving notably higher relevance and correctness than MMSearch‑R1 and Graph‑R1. It also excels in comprehensiveness, factuality, logical coherence, and knowledgeability, demonstrating that self‑evolving graph refinement reduces hallucination while preserving reasoning quality. Moreover, EvoGraph‑R1 attains the highest diversity score, confirming that \textsc{WebSearch} integration and dynamic graph expansion effectively incorporate varied evidence sources.

\subsection{Performance in Low-Resource Settings - RQ5}

\begin{figure}[htbp]
    \centering
    \includegraphics[width=0.92\linewidth]{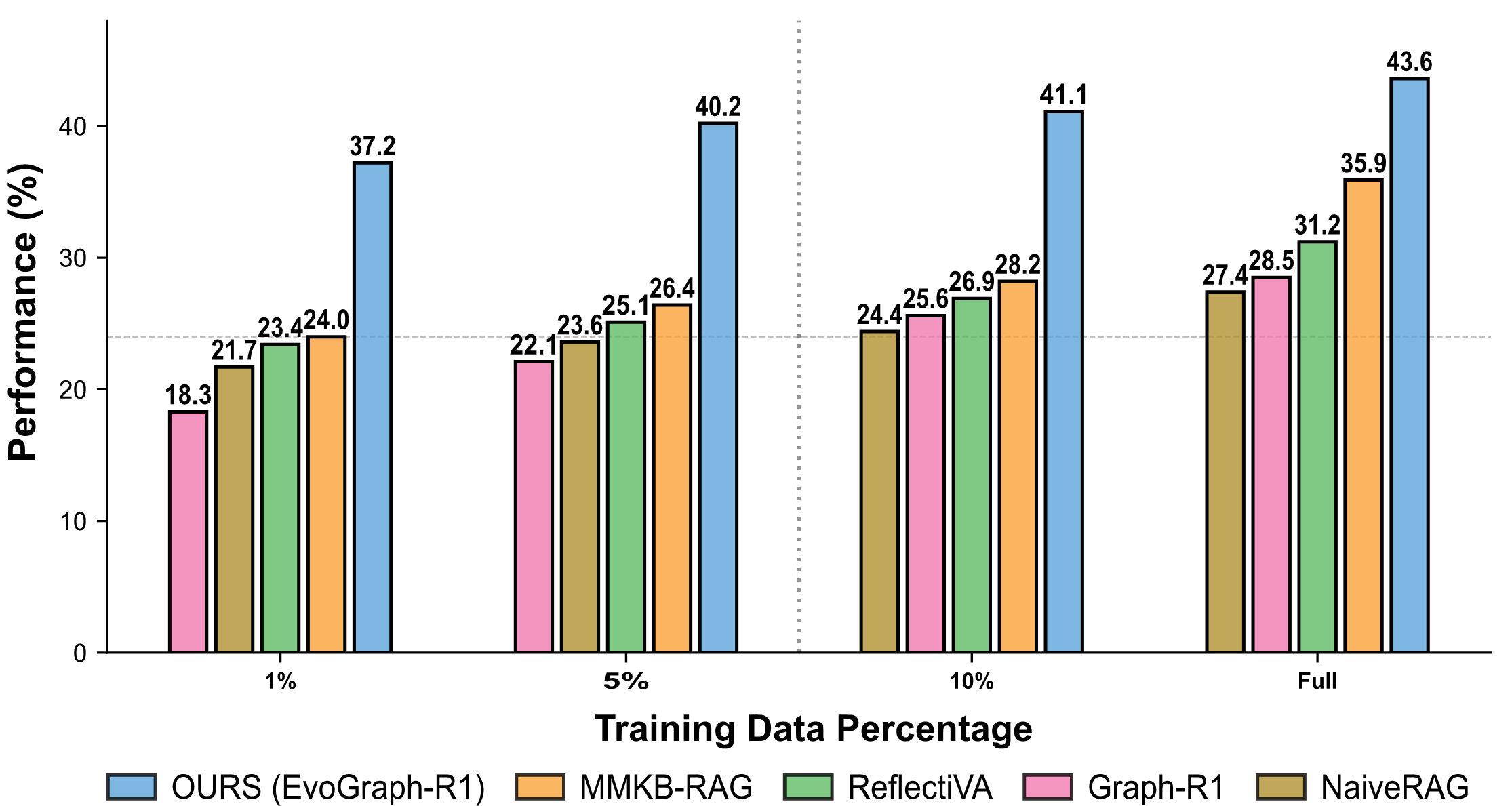}
    \caption{\textbf{Performance under limited corpus size on E-VQA.} EvoGraph-R1 (blue) consistently outperforms all baselines, with the performance gap widening as corpus size decreases.}
    \label{fig:low_resource}
\end{figure}

We evaluate robustness on E-VQA by restricting Wikipedia to 1\%, 5\%, and 10\% of its full size, simulating limited-resource settings. Figure~\ref{fig:low_resource} shows resilience. At 1\% corpus, EvoGraph-R1 achieves 37.2\% accuracy and surpasses baselines ranging from 13.2\% to 18.9\%. Notably, the performance gap widens as data decreases. Compared to MMKB-RAG, the gain increases from +7.7 points at full corpus to +13.2 points at 1\% corpus, and remains large at 5\% with +13.8 and at 10\% with +12.9. This trend underscores the importance of \textsc{WebSearch} and dynamic \textsc{INSERT} when static knowledge is insufficient.

\subsection{Graph Refinement Analysis}
\begin{figure}[h!]
	\centering
	\subfloat[Before refinement]{
		\includegraphics[width=0.4\linewidth,height=4cm,keepaspectratio]{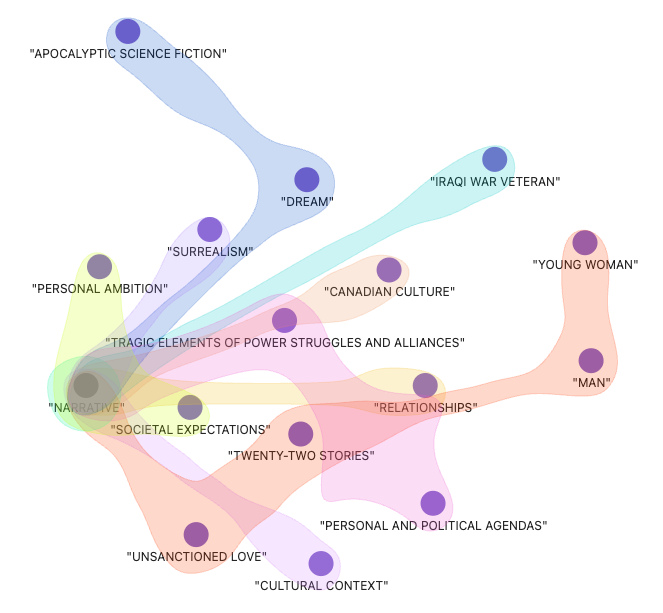}
		\label{fig_sub1}
	}
	\subfloat[After refinement]{
		\includegraphics[width=0.48\linewidth,height=4cm,keepaspectratio]{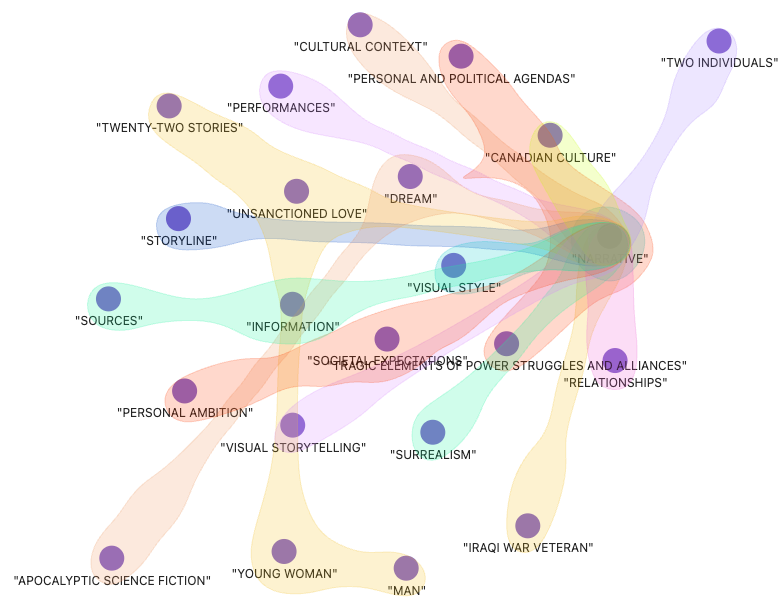}
		\label{fig_sub2}
	}
	\caption{\textbf{Visualization of the hypergraph centered on ``NARRATIVE'' before and after refinement.} The refined graph exhibits clearer thematic structure and deeper reasoning paths.}
	\label{fig:narrative_graph_refine}
    \vspace{-3mm}
\end{figure}

We visualize the subgraph centered on “NARRATIVE” before and after refinement (Figure~\ref{fig:narrative_graph_refine}). The pre-refinement graph is shallow and loosely organized, with fragmented single-hop associations and weak thematic cohesion. After refinement, the graph consolidates into denser and more coherent clusters, revealing clearer multi-hop thematic connections and a more structured narrative space.

For quantitative evaluation, we employ standard complex-network metrics \cite{newman2003structure} to measure entity and edge counts, graph density, and the Watts–Strogatz clustering coefficient \cite{watts1998collective}, capturing structural cohesion. To assess semantic consistency, we compute edge semantic similarity using Sentence-BERT embeddings \cite{reimers2019sentence}. As summarized in Table~\ref{tab:graph_refinement_stats}, refinement yields consistent improvements in all structural and semantic metrics, indicating a more connected, coherent, and aligned graph.

    \begin{table}[t]
    \centering
    \footnotesize
    \caption{\textbf{Graph refinement statistics before and after refinement.}
    Percentage change is computed as $(\text{After} - \text{Before}) / \text{Before}$.}
    \label{tab:graph_refinement_stats}
    \begin{tabular}{lccc}
    \toprule
    \textbf{Metric} & \textbf{Before} & \textbf{After} & \textbf{$\Delta\%$} \\
    \midrule
    Nodes & 120{,}499 & 123{,}631 & +2.60\% \\
    Hyperedges & 177{,}408 & 181{,}418 & +2.26\% \\
    Graph Density & 0.781 & 0.842 & +7.81\% \\
    Clustering Coefficient & 0.024 & 0.028 & +16.67\% \\
    Edge Semantic Similarity & 0.664 & 0.685 & +3.16\% \\
    \bottomrule
    \end{tabular}
    \vspace{-3mm}
    \end{table}

\section{Conclusion}
We introduced EvoGraph-R1, a self-evolving GraphRAG framework that reconceptualizes knowledge graphs as dynamic MDP environments rather than static structures. An autonomous agent iteratively refines a multimodal knowledge hypergraph, transforming sparse initial graphs into coherent, query-specific knowledge states. Comprehensive experiments demonstrate state-of-the-art performance on both text and multimodal benchmarks, validating that self-evolving graphs generalize across modalities. This work establishes agent-driven graph evolution as a fundamental paradigm for adaptive retrieval, enabling robust reasoning over incomplete knowledge and dynamic environments.

\newpage
{
    \small
    \bibliographystyle{ieeenat_fullname}
    \bibliography{main}
}




\end{document}